\documentclass[journal]{IEEEtran}
\usepackage{graphicx}
\usepackage{tikz}
\usepackage{comment}
\usepackage{color}
\usepackage{graphicx}
\usepackage{booktabs}
\usepackage{multirow}
\usepackage{epsfig}
\usepackage[normalem]{ulem}
\usepackage{amsmath, bm, subfigure, epstopdf, url, pifont, overpic, cases}
\usepackage{latexsym, amssymb, bbding, multirow, makecell,  diagbox, enumitem}
\usepackage{array, enumitem, soul, algorithm, algpseudocode}
\usepackage{xcolor}
\usepackage{hyperref}
\usepackage[capitalize]{cleveref}
\usepackage{colortbl}

\Crefname{section}{Section}{Sections}
\Crefname{table}{Table}{Tables}
\Crefname{figure}{Figure}{Figures}
\definecolor{mycyan}{cmyk}{.3,0,0,0}
\RequirePackage[normalem]{ulem} %DIF PREAMBLE
\RequirePackage{color}\definecolor{RED}{rgb}{1,0,0}\definecolor{BLUE}{rgb}{0,0,1}

\begin{document}

\title{A Practical Contrastive Learning Framework for Single Image Super-Resolution}

\author{Gang~Wu,
        Junjun~Jiang,~\IEEEmembership{Senior Member,~IEEE,~}
        and Xianming Liu,~\IEEEmembership{Member,~IEEE}

\IEEEcompsocitemizethanks{
\IEEEcompsocthanksitem G. Wu, J. Jiang, and X. Liu are with the School of Computer Science and Technology, Harbin Institute of Technology, Harbin 150001, China. E-mail: \{gwu, jiangjunjun, csxm\}@hit.edu.cn (Corresponding author: Junjun Jiang).

}
}
\markboth{Journal of \LaTeX\ Class Files,~Vol.~14, No.~8, August~2021}%
{Shell \MakeLowercase{\textit{et al.}}: A Sample Article Using IEEEtran.cls for IEEE Journals}

\maketitle
\begin{abstract}
Contrastive learning has achieved remarkable success on various high-level tasks, but there are fewer contrastive learning-based methods proposed for low-level tasks. It is challenging to adopt vanilla contrastive learning technologies proposed for high-level visual tasks to low-level image restoration problems straightly. Because the acquired high-level global visual representations are insufficient for low-level tasks requiring rich texture and context information. In this paper, we investigate the contrastive learning-based single image super-resolution from two perspectives: positive and negative sample construction and feature embedding. The existing methods take naive sample construction approaches (e.g., considering the low-quality input as a negative sample and the ground truth as a positive sample) and adopt a prior model (e.g., pre-trained VGG \cite{VGG} model) to obtain the feature embedding. To this end, we propose a practical contrastive learning framework for SISR, named PCL-SR. We involve the generation of many informative positive and hard negative samples in frequency space. Instead of utilizing an additional pre-trained network, we design a simple but effective embedding network inherited from the discriminator network which is more task-friendly. Compared with existing benchmark methods, we re-train them by our proposed PCL-SR framework and achieve superior performance. Extensive experiments have been conducted to show the effectiveness and technical contributions of our proposed PCL-SR thorough ablation studies. The code and pre-trained models can be found at \url{https://github.com/Aitical/PCL-SISR}.

\end{abstract}

%%%%%%%%% BODY TEXT
\section{Introduction\label{section:introduction}}
Recently, contrastive learning has been an effective paradigm for unsupervised representation learning. Most approaches are developed based on the instance discrimination assumption and learn visual representations by making views from the same instance similar and that from different instances dissimilar \cite{instance_dis,SimCLR,MoCo,MoCov2,SimSiam}. The learned semantic representations are beneficial to a wide variety of downstream tasks, particularly high-level tasks, and achieve promising results, e.g., supervised image classification \cite{SupCL}, image clustering \cite{contrastive_clustering}, fine-gained image classification \cite{contrastive_fine_gained}, and knowledge distillation \cite{contrastive_distillation}. 

When it comes to low-level image processing tasks, there are some challenges to directly applying contrastive learning approaches. Firstly, the learned global visual representations are inadequate for low-level tasks that call for rich texture and context information. Secondly, a series of data augmentations have been developed to generate positive and negative samples for high-level downstream tasks \cite{SimCLR,MoCo,what_makes_good,joint}. However, except for some simple geometric augmentations (e.g., rotation augmentation in the widely used self-ensemble scheme), most of the complicated data augmentations cannot maintain the dense pixel correspondences and thus are not suitable for low-level tasks. Thirdly, a meaningful latent space (or embedding space) is required for contrastive loss. In contrast to high-level tasks that try to obtain the best semantic representations, low-level tasks aim at reconstructing restored results in data space. It is of great importance to explore a proper and meaningful embedding space where the contrastive loss can be effectively defined.

Current contrastive learning-based methods for low-level tasks mainly focus on exploiting negative samples, while let the ground truth image as the positive sample. For example, \cite{contrastive_dehazing} treats the degraded image (the input hazy image) as a negative sample and presents a novel image dehazing method with a contrastive regularization. In \cite{wang2021towards} and \cite{han2021single}, they take other examples in the dataset as negative samples for contrastive image super-resolution and underwater image restoration. These methods make a preliminary attempt and have proved the effectiveness of incorporating the contrastive constraint to low-level tasks. Another line of research is to model the statistical characteristics of the image by contrastive learning. Dong \emph{et al}. \cite{dong2021residual} assume that two patches from the same sample have similar noise distributions and two these from two different samples have two different noise distributions and develop a residual contrastive loss for joint demosaicking and denoising. In a similar vein, Chen \emph{et al}. \cite{chen2021unpaired} develop an unpaired image deraining method based on rain space contrastive learning, which can better help rain removal when compared with the original image space. In \cite{NIPS_CSR}, Zhang \emph{et al}. present a contrastive learning strategy in the feature channel space to obtain resolution-invariant features. They take feature maps of different channels as samples and postulate that the corresponding channel of low-resolution (LR) and high-resolution (HR) feature maps are positive while those from different are negative. Wang \emph{et al} \cite{DASR} apply the contrastive loss to pre-train a kernel estimation model which aims at separating different degradations and obtaining the degradation-aware representation.

\cref{tab:taskcomp} summarizes the characteristics of current contrastive learning-based image restoration methods, which are presented most recently. The positive samples are defined as the ground truth, while the negative samples are simply defined as the degraded images or other images in the dataset \cite{contrastive_dehazing,wang2021towards,han2021single}. These negative samples are dissimilar to the reconstructed image and easily distinguished, \emph{i.e.}, they are too distant to contribute to the contrastive loss. According to the specific image restoration tasks, another line of research \cite{dong2021residual,chen2021unpaired,NIPS_CSR,DASR} try to generate some invariant (global) features of image, which are immune to the noise, rain, resolution, and blur, based on constrictive loss. They overlooked the ingredient of constructing effective positive and negative pairs for the reconstructed image. In addition, since the constrictive losses of these methods are defined on some specific embedding space and cannot well generalize to other methods.

In this paper, we propose a practical contrastive learning framework named PCL-SR and we investigate it from two perspectives: sample construction and feature embedding. As revealed by recent studies, super-resolved results of current deep learning methods are smooth and look implausible (may be averaged from all possible outputs of the SR network). Based on these observations and hard negative mining studies \cite{hard_c1,hard_c2}, we propose to generate multiple hard negative samples by applying some slight blurry to the ground truth and generate multiple positive samples by simply sharpening the ground truth, resulting in informative positive and hard negative pairs for the super-resolved image. In this way, we believe that more hard negative samples will encourage the super-resolved image far from smooth results, while more positive samples will force the network to draw in more detailed information. 
On the other hand, in contrast to adopting a prior model as the embedding network (e.g., pre-trained VGG \cite{VGG} network), we propose to leverage a cheap and task-friendly feature embedding network, the discriminator of SR network, to embed positive/negative/anchor samples into a proper feature space where the contrastive loss will be effectively defined. 

Our contributions are summarized as follows: (i) We propose a practical and general contrastive learning framework for the SISR task. We take a valid data augmentation strategy to generate informative positive and negative samples for SISR. (ii) We rethink and explore a novel way to obtain a task-friendly embedding network where contrastive loss works efficiently by reusing the discriminator of SR network. (iii) Extensive experiments show that our method, dubbed \textit{PCL-SR}, outperforms several representative SISR methods. In addition, ablation studies are conducted to analyze different components of our proposed method.

In the following section we will first give some related work of single image super-resolution and contrastive learning approaches in \cref{sec:related_work}. In \cref{sec:related_work}, we introduce our proposed PCL framework and explain it in detail. Then, \cref{sec:experiments} describes our training settings and experimental results including ablation analysis, where we compare the performance of our approach to other state-of-the-art methods. Finally, some conclusions are drawn in \cref{sec:conclusion}.

\begin{table*}[t]
\centering
  \caption{Comparison {between} the proposed contrastive learning framework {and} current contrastive learning approaches for low-level tasks.  \label{tab:taskcomp}}
\renewcommand{\arraystretch}{1.2}

 \resizebox{0.975\textwidth}{!}{%
  \begin{tabular}{|c|c|c|c|}
    \hline
  %  \centering
    Ref.     & Task      &{Positive$/$Negative Samples}  & {Feature Embedding Space}  \\
    \hline
    \hline
    {CVPR'21 {\cite{contrastive_dehazing}}} &{Dehazing} & {Ground Truth$/$Hazy Image} &  {Additional VGG Embedding}     \\
    {IJCAI'21 {\cite{wang2021towards}}} &{SISR} & {Ground Truth$/$Other Images} &  {Additional VGG Embedding}     \\
    {IGARSS'21 {\cite{han2021single}}} &{Underwater Image Restoration} & {Ground Truth$/$Degraded Image}  &  {Additional VGG Embedding} \\
    {TIP'22} {\cite{TACL}} &{Underwater Image Restoration} &    {Ground Truth$/$Degraded Image}  &     {Additional VGG Embedding} \\
    {arXiv'21 {\cite{NIPS_CSR}}} &{Blind SR} & {Same Channel Feature$/$Different Channel Features} &  {Task-specific Embedding}   \\
    {CVPR'21 {\cite{DASR}}} &{Blind SR} & {Same Blur Style$/$Different Blur Styles} &  {Task-specific Embedding}   \\
   {CVPR'22 {\cite{chen2021unpaired}}} &{Unpaired Deraining} & {Same Rain Style$/$Different Rain Styles} &  {Task-specific Embedding} \\
   {arXiv'21 {\cite{dong2021residual}}} &{Demosaicking and Denoising} & {Same Noise Style$/$Different Noise Styles}  &  {Task-specific Embedding} \\
    %\midrule
    {\textbf{Ours}} &{\textbf{SISR}} & {\textbf{GT and Multiple Sharpen Images$/$Multiple Slightly Blurry Images}} &  {\textbf{Task-Generalizable Embedding}} \\
    \hline
  \end{tabular}
  }
\end{table*}

\section{Related work \label{sec:related_work}}
In this section, we will briefly introduce the related work about self-supervised contrastive learning methods, contrastive learning methods for image restoration, contrastive learning for image-to-image translation, and single image super-resolution methods.

\subsection{Contrastive Learning}
 Contrastive learning has emerged as an effective paradigm for unsupervised representation learning, by maximizing mutual information. In recent years, several works \cite{CPC,infoMax,instance_dis,SimCLR,MoCo} have studied this paradigm. The contrastive loss, which is similar to previous works \cite{contrastive_lecun,triplet,N-pair} in deep metric learning, aims to push the anchor sample away from negative samples and pull it closer to positive samples in the latent space. The selection of negative and positive samples is crucial and depends on downstream tasks. In \cite{instance_dis,SimCLR,MoCo}, randomly augmented samples from the anchor sample serve as positive samples, while samples from others are negative. In \cite{what_makes_good}, Tian \textit{et al.} analyze the optimal augmentation strategy in detail. In \cite{joint}, Cai \textit{et al.} propose a feature space augmentation strategy, sampling numerous positive and negative samples from the latent feature distribution. In \cite{hard_c1,hard_c2,neg_aug}, authors analyzed the impact of hard negative samples and exploited hard negative mining strategies to dig more informative negative samples.

\subsection{Constrictive Learning for Image Restoration}
For single image dehazing, Wu \textit{et al.} \cite{contrastive_dehazing} propose a method that utilizes a pre-trained VGG model to obtain latent embeddings, where positive and negative samples are selected from the ground truth and corresponding hazy image, respectively, and contrastive loss is conducted using intermediate feature maps extracted from the VGG network. For blind super-resolution (BSR), Wang \textit{et al.} \cite{DASR} apply the contrastive loss to pre-train a kernel estimation model, which obtains a degradation-aware representation by distinguishing different degradations. The authors assumed that patches from the same image are under the same degradation and that patches from different images are not the same, thus selecting positive and negative samples accordingly. Zhang \textit{et al.} \cite{NIPS_CSR} propose a two-stage BSR method. Firstly, an image encoder is pre-trained to learn resolution-invariant features by contrastive loss. Then, another contrastive loss is adopted to finetune the super-resolved results.

{Recently, Liu \textit{et al.} \cite{TACL} introduce the contrastive prior into underwater image enhancement, which takes observed underwater images as negative samples and clear in-air images as positive and the pre-trained VGG network is adopted as the feature embedding network to provide multi-layer intermediate features.}

\subsection{Constrictive Learning for I2I Translation}
CUT \cite{cut} is a pioneering work that proposed patch-based contrastive loss for unpaired image-to-image (I2I) translation, which effectively obtains structure-preserved and style-transferred features in latent space. For a given patch of the input image, positive and negative samples are the corresponding transferred patch and other random patches from the same domain. Han \textit{et al.} \cite{dclgan} propose the DCLGAN, which utilizes a bidirectional patch-based contrastive loss between the source and target domain.{ In recent work, Lin \textit{et al.} \cite{RankNCE} explore the effectiveness of negative samples in patch-based contrastive loss and proposed a negative pruning technology called RankNCE, which reduces the number of negative samples by ranking the similarity score between the corresponding negative patches and the anchor patch feature. Ko \textit{et al.} \cite{Ko_2022_CVPR} introduce the self-supervised paradigm proposed in \cite{SimSiam} into the I2I task and explored a self-supervised dense consistency regularization (DCR), where two augmented samples with an overlap region are taken as the input, and the DCR is calculated on the dense representation of the overlap region.}

\subsection{Single Image Super-Resolution}
Deep-learning-based methods have dominated the single image super-resolution field in recent years. Dong \emph{et al}. propose the first CNN-based SR method, named SRCNN. Since that, various efficient and deeper architectures have been proposed for SR and the performance on benchmark datasets has been continuously improved by newly developed network architectures \cite{EDSR,RDN,DBPN,RCAN,tang2019deep,xin2020wavelet}. In addition, Zhou \textit{et al.} \cite{IGNN} propose adaptive patch feature aggregation with a graph neural network named IGNN. Chen \textit{at al.} \cite{IPT} proposed a multi-task pre-trained image processing transformer named IPT. Liang \textit{et al.} \cite{SwinIR} develop the SwinIR method, which introduces a hybrid architecture that combines Swin-Transformer \cite{Swin-Trans} and convolutions into images restoration tasks. Besides investigating more powerful network architectures, some perceptual-driven approaches also have been proposed utilizing perceptual loss functions to achieve better visual quality \cite{perceptual_SR,SRGAN,ESRGAN,RankSRGAN}.

{In this paper, we present a practical contrastive learning framework for single image super-resolution named PCL. We investigate the contrastive learning paradigm for SISR from two perspectives: sample generation and feature embedding. In contrast to existing methods, which propose hard negative samples from semantic discrimination \cite{hard_c1,hard_c2,neg_aug} or patch feature similarity \cite{RankNCE}, we exploit task-valid hard negative samples for SISR tasks by performing a slight blur operation. Moreover, while most existing methods adopt pre-trained VGG as the feature embedding network, we propose an adaptive frequency-aware feature embedding approach, which learns task-friendly embedded features and is able to be sensitive to high-frequency information.}

\section{Method\label{sec:methods}}
\begin{figure*}
    \centering
    \includegraphics[width=0.995\textwidth]{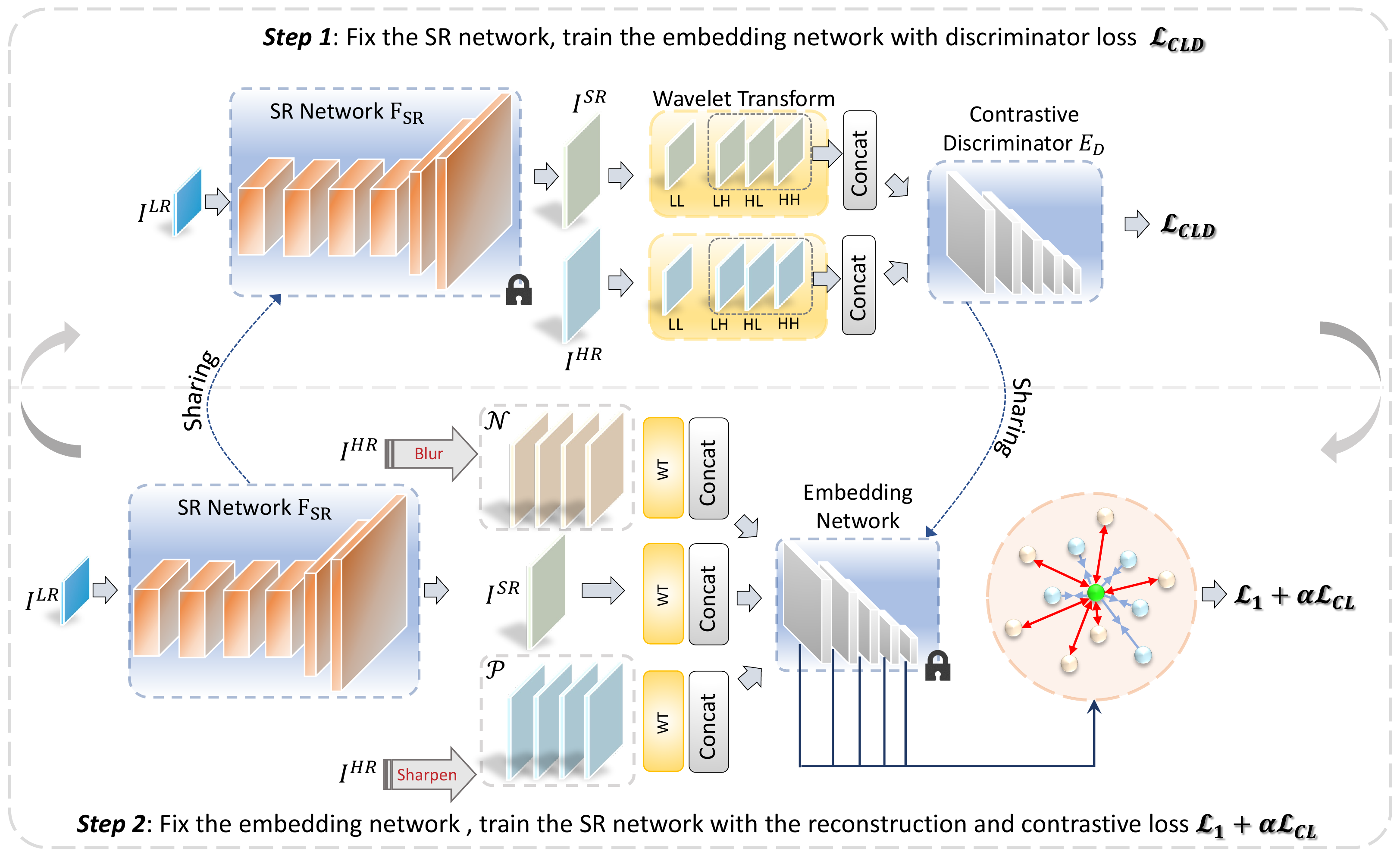}
    \caption{Overview of our proposed contrastive learning framework for SISR. We adopt the GAN-like training processing that updating our embedding network $E_{D}$ and target SR network $F_{SR}$ iteratively. We train our embedding network to learn degradation-aware features. Then $E_{D}$ is frozen and SR network is trained with the pixel-wise construction loss $\mathcal{L}_P$ and our contrastive loss $\mathcal{L}_{CL}$.}
    \label{fig:framework}
\end{figure*}
In this section, we introduce our proposed practical contrastive learning framework for SISR in detail. We first introduce the preliminaries of contrastive learning and then we describe our positive and negative sample generation strategy and the training of our feature encoder. At last, we present our main framework which employs contrastive learning to further improve the performance of existing SISR works.

\subsection{Preliminaries}\label{section:preliminaries}
Contrastive learning is one of the most powerful approaches for representation learning. It aims at pulling the anchor sample close to the positive samples and pushing it far away from negative samples in latent space \cite{instance_dis, SimCLR, MoCo}. For the image dataset $\mathcal{I}$, the representation learning model $E$ is trained to extract representations $\mathcal{R}=\{r_i|r_i = E(I_i), I_i \in \mathcal{I}\}$ with InfoNCE loss \cite{CPC, infoMax}. The loss $\mathcal{L}_{InfoNCE}$ is based on a Softmax formulation and for the $i$-th sample the loss $\mathcal{L}_i$ is as formulated as follows:
\begin{equation}
    \mathcal{L}_i=-\log \frac{{\rm exp}(r_i^{T}\cdot r_i^{+}/\tau)}{{\rm exp}(r_i^{T}\cdot r_i^{+}/\tau) +\sum_{j=1}^{K}{\rm exp}(r_i^{T}\cdot r_j^{-}/\tau)},
\end{equation}
where $\tau$ is the temperature hyper-parameter. $r_i^{+}$ means the representation of the positive sample usually generated by random data augmentations from the same sample $I_i$. $K$ is the number of negative samples and $\{r_j^{-}\}_{j=1}^{K}$ is the set of negative representations from negative samples $\{I_j| I_j\in \mathcal{I}, j\neq i\}_{j=1}^{K}$ that are random selected other images from dataset. Finally, total contrastive loss is as follows:
\begin{equation}
    \mathcal{L}_{InfoNCE}=\frac{1}{N}\sum_{i=1}^{N}\mathcal{L}_i.
\end{equation}
In addition, the work in \cite{SupCL} modified and applied contrastive loss to the supervised classification task where there are more than one positive samples. For the $i$-th image, this supervised contrastive loss is as follows:

\begin{equation}
    \mathcal{L}_{i}=- \frac{1}{P}\sum_{p=1}^{P}\log\frac{{\rm exp}(r_i^{T}\cdot r_{p}^{+}/\tau)}{{\rm exp}(r_i^{T}\cdot r_i^{+}/\tau) +\sum_{j=1}^{K}{\rm exp}(r_i^{T}\cdot r_j^{-}/\tau)}
    \label{eq:multi_cl},
\end{equation}
where $P$ is the number of the positive set, noted as $\{r_p^+\}_{p=1}^{P}$. Then the total supervised contrastive loss $\mathcal{L}_{SupCL}$  is:
\begin{equation}
    \mathcal{L}_{SupCL}=\frac{1}{N}\sum_{i=1}^{N}\mathcal{L}_i.
\end{equation}

Contrastive learning is beneficial for various downstream tasks and achieves promising performance. As we described in \cref{section:introduction},  we can find that to employ contrastive learning methods, carefully designed sample selection and construction strategy and a task-related latent space needs to be explored.
Next, we will describe our sample generation strategy from the perspective of frequency domain and how to train a task-friendly embedding network in SISR instead of a pre-trained model, such as VGG network.

\subsection{Positive and Negative Sample Generations}\label{section:selection}
The SISR task aims to transform low-resolution images (noted as $\mathcal{I}^{LR}$) into sharp, realistic, and high-resolution images (noted as $\mathcal{I}^{HR}$). As LR images are formulated from an image degradation process and contain only the low-frequency information, SR model focuses on learning a reverse translation to recover the lost high-frequency components (e.g., edge and texture information). With this in mind, we take valid data augmentations to generate our positive and negative samples. 

\textbf{Informative positive sample generation}. In addition to the only HR ground truth, we further generate $K_P$ sharpened images as positive set $\mathcal{P}_i$ by applying different high-pass kernels on HR image. 
For the $i$-th image, we denote its positive set as follows:
\begin{equation}
    \begin{aligned}
    \mathcal{P}_i= &\{P_j|P_j = {\rm Sharpen}(I^{HR}_i) \}_{j=1}^{K_{P}},
    \end{aligned}
\end{equation}
where $K_P$ is the number of positive samples. ${\rm Sharpen}$ presents a random sharpness function. This is different from the existing contrastive learning-based image restoration methods, which consider only the ground truth as the positive sample \cite{NIPS_CSR,contrastive_dehazing}. It should be noted that in order to generate more informative positive samples, we apply different high-pass kernels to the HR ground truth. This positive sample generation strategy is designed following two observations: (i) the object of the SISR task is to obtain detailed results. We can use some informative positive samples to induce more high-frequency details for the reconstruction results. (ii) SISR is an ill-posed problem, and the mapping between the LR and HR images is ``one-to-many''. That is, the number of ground truth should not only be one, and there are many possible HR samples except the given ground truth \cite{jo2021tackling}. Our proposed positive sample generation method can be seen as a very coarse one. 

\textbf{Hard negative sample generation}. In order to introduce contrastive learning to the low-level image restoration problems, recent works \cite{contrastive_dehazing,wang2021towards,han2021single} simply take degraded images (e.g. the input hazy image or LR image) or other images in the dataset. When compared with the super-resolved image, low-quality negative samples are dissimilar and easy to be distinguished. Inspired from the hard negative samples mining and adversarial training methods \cite{hard_c1,hard_c2,AdCo}, a natural idea is whether we can feed some difficult hard examples that are similar to ground truth as hard negative samples. 
 
Here we generate slightly blurry images from the ground truth as our hard negative sample set $\mathcal{N}_i$ because they are close to the ground truth, thus forcing the reconstructed SR image become closer to the ground truth. For the $i$-th image, we denote its negative set as follows:
\begin{equation}
    \begin{aligned}
    \mathcal{N}_{i}=&\{N_j|N_j={\rm Blur}(I^{HR}_i)\}_{j=1}^{K_N},
    \end{aligned}
\end{equation}
where $K_N$ is the number of negative samples and we use $K_N=K_P=4$ as default. ${\rm Blur}$ is blur function with {a} random Gaussian kernel. {For the blur kernel setting, the size is uniformly sampled from $\{3, 5, 7, 9, 11\}$, and random Gaussian kernel samples the kernel width uniformly from $[0.3, 1.5]$.}

\subsection{Feature Embedding Network}\label{section:embedding}
In this section, we introduce a simple but efficient way to obtain a task-friendly embedding network. As described in \cref{section:introduction}, VGG based perceptual loss is widely adopted \cite{SRGAN,ESRGAN} and recent work in \cite{contrastive_dehazing} designs a contrastive loss based on the pre-trained VGG model. We believe that a task-friendly embedding network is better because features obtained by pre-trained VGG tend to be high-level semantic information. In addition, compared with the SISR task, pre-training with ImageNet is a very heavy task. Furthermore, a good embedding network for SISR should be degradation-aware in order that contrastive loss can work even if the super-resolved results are very close to the ground truth. In other words, a good embedding network should be able to distinguish changes in detail. 

Inspired by adversarial learning approaches in \cite{GAN,SRGAN}, we find that the discriminator learned in the vanilla GAN framework is degradation-aware because it can correctly distinguish whether the input image is fake or not. With this in mind, we employ a GAN-like framework to obtain our task-friendly embedding network by forcing it to distinguish SR and HR images as illustrated in \cref{fig:framework}. Notably, to enhance the high-frequency components learning, we separate the image to low- and high-frequency parts and our embedding network is trained with only the frequency components, which has been verified to be effective in real-world SR problems \cite{fritsche2019frequency}. Here we use Haar wavelet transform to extract the informative high-frequency components. The four sub-bands decomposed by Haar wavelet transform, noted as $LL$, $LH$, $HL$, and $HH$. Then we stack the three high-frequency-related components ($LH$, $HL$, and $HH$) as the input and feed them to discriminator network $E_{D}$. Instead of training with conventional real or fake binary classification network, we adopt a contrastive discriminator loss proposed in \cite{ContrasD} to train our $E_{D}$. It is a one-against-a-batch classification in the Softmax cross-entropy formulation, and thus the loss function of the $E_{D}$ can be formulated as follows:

\begin{equation}
\begin{aligned}
\mathcal{L}_{CLD} &= \mathbb{E}_{I^{H R}}\left[\log \frac{e^{E_{D}\left(\mathcal{H}_{w}\left(I^{H R}\right)\right)}}{e^{E_{D}\left(I^{H R}\right)}+\sum_{I^{L R}} e^{E_{D}\left(\mathcal{H}_{w}\left(I^{SR}\right)\right)}}\right] \\
&+\mathbb{E}_{I^{LR}}\left[\log \frac{e^{-E_{D}(\mathcal{H}_{w}(I^{SR}))}}{e^{-E_{D}(I^{SR})}+\sum_{I^{HR}} e^{-E_{D}(\mathcal{H}_{w}(I^{HR}))}}\right],
\end{aligned}
\end{equation}
where $I^{SR} = F_{SR}(I^{LR})$ represents super-resolved image, and $\mathcal{H}_{w}$ is the operator extracting $LH$, $HL$ and $HH$ subbands and concatenating them.

\subsection{Contrastive Loss}
As described in \cref{section:selection} and \cref{section:embedding}, we employ a task-friendly embedding network to obtain a frequency-based latent space, and we also generate sharpness and blurry images as our multiple positive and negative samples. To fully utilize these samples, we conduct our contrastive loss with multi-intermediate features from embedding network $E_{D}$ and propose our contrastive loss based on the \cref{eq:multi_cl}. 
For the target super-resolved image $I^{SR}_i$, its generated positive and negative sets are noted as $\mathcal{P}_i$ and $\mathcal{N}_i$ respectively. The feature representations for the super-resolved image, positive sample, and negative sample are noted as $f$, $p$, and $n$, respectively. We use superscript $l$ as the layer index in $E_{D}$. Our contrastive loss for the $i$-th sample on the $l$-th layer is defined as follows:

\begin{small}
\begin{equation}
\begin{aligned}
    \mathcal{L}_{i,l}= \frac{1}{K_P}\sum_{j=1}^{K_P}- \log\frac{{\rm exp}(s(f_{i}^{l},p_{j}^{l})/\tau)}{{\rm exp}(s(f_{i}^{l}, p_{j}^{l})/\tau) +\sum_{k=1}^{K_N}{\rm exp}(s(f_{i}^{l},n_{k}^{l})/\tau)},
\end{aligned}
    \label{eq:our_cl}
\end{equation}
\end{small}where $s$ is the similarity function. Let's note the shape of feature map as $C\times H \times W$. We adopt the mean value of pixel-wise cosine similarity as the similarity between feature maps. $s$ is defined as follows:

\begin{equation}
    \begin{aligned}
    s(f^x, f^y) = \frac{1}{HW}\sum_{h=1}^{H}\sum_{w=1}^{W}\frac{f^x_{hw}f^{y}_{hw}}{\|f^x_{hw}\|\|f^{y}_{hw}\|}
    \end{aligned}.
\end{equation}

Then, our total contrastive loss $\mathcal{L}_{CL}$ is as follows:

\begin{equation}
\begin{aligned}
    \mathcal{L}_{CL}= \frac{1}{NL}\sum_{i=1}^{N}\sum_{l=1}^{L}\mathcal{L}_{i,l},
\end{aligned}
    \label{eq:final_our_cl}
\end{equation}
where $N$ is the number of the training images and $L$ is the number of feature layers we used. We take the first 4 intermediate layers to calculate our contrastive loss $\mathcal{L}_{CL}$ and the $L$ is 4 as default.

\subsection{Training and Implementation Details}
For image reconstruction, we utilize $\ell_{1}$ loss as follows:
\begin{equation}
    \mathcal{L}_1=\frac{1}{N}\|I^{SR}_{i}-I^{HR}_{i}\|_1
\end{equation}
We train our SR network $F_{SR}$ using $L_1$ loss and the proposed contrastive loss in \cref{eq:final_our_cl}. The total loss function $\mathcal{L}_{SR}$ is defined as follows:
\begin{equation}
    \mathcal{L}_{SR} = \mathcal{L}_1+\alpha\mathcal{L}_{CL},
\end{equation}
where $\alpha$ is a scaling parameter and we take $\alpha=${0.1} as default.

\begin{figure*}
    \centering
    \includegraphics[width=0.99\textwidth]{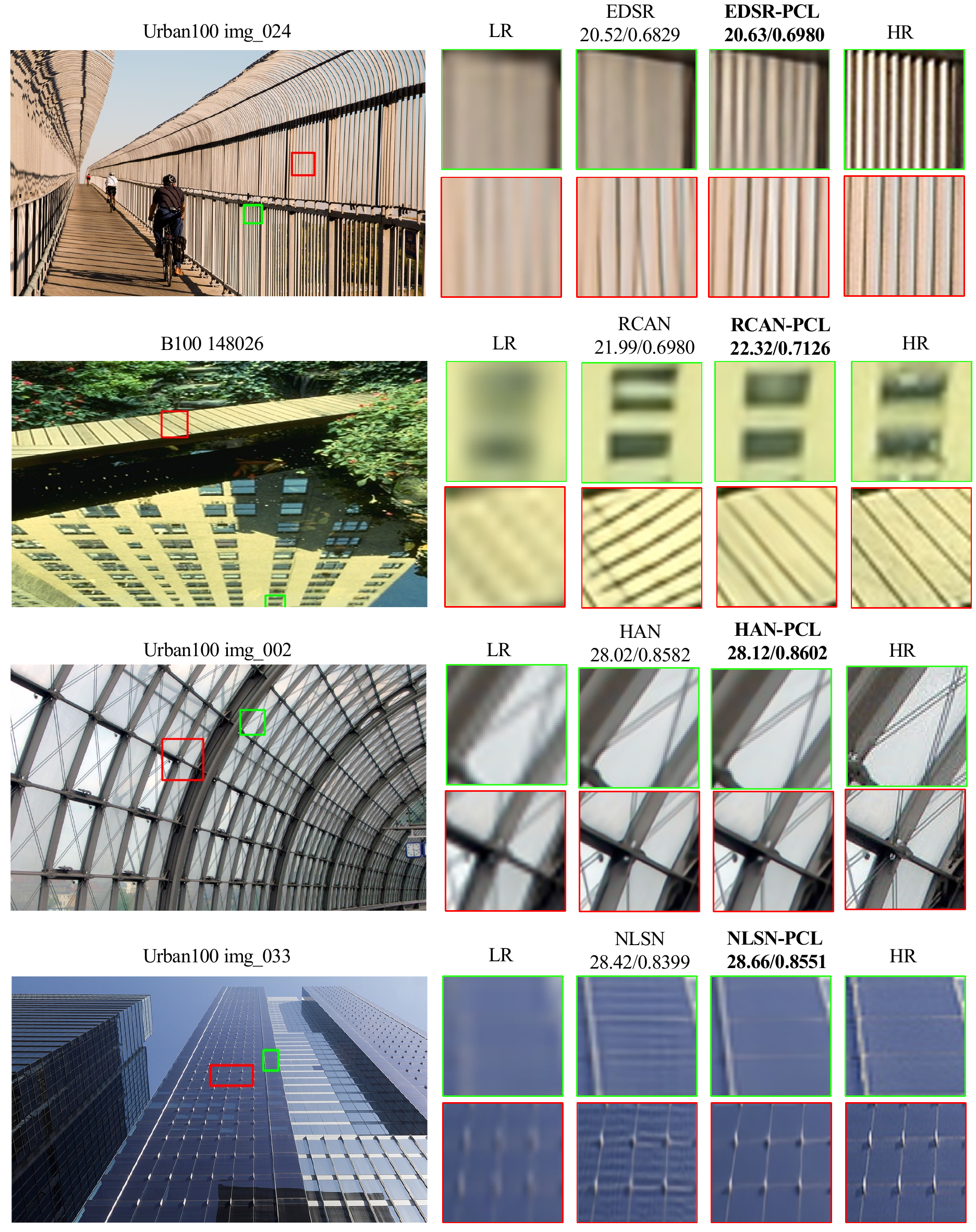}
    \caption{{\textbf{Visual comparison} between the results of benchmark methods and our re-trained counterparts. PSNR and SSIM scores are presented.}}
    \label{fig:sr_visual}
\end{figure*}

\begin{table*}[tbp]
\begin{center}
\caption{Quantitative comparison with SoTA methods on benchmark datasets (PSNR (dB)/SSIM). Results of our PCL are in \textbf{bold} and the improvements are in \textcolor{green}{green}.}
\label{tab:main_psnr_ssim}
\renewcommand{\arraystretch}{1.1}
\resizebox{0.8\textwidth}{!}{%
\begin{tabular}{|c|c|c|c|c|c|c|c|c|c|c|}
\hline
\multirow{2}{*}{Method}     & \multicolumn{2}{c|}{Set5} & \multicolumn{2}{c|}{Set14} & \multicolumn{2}{c|}{B100} & \multicolumn{2}{c|}{Urban100} & \multicolumn{2}{c|}{Manga109}  \\
\cline{2-11}
    & PSNR  & SSIM             & PSNR  & SSIM              & PSNR  & SSIM             & PSNR  & SSIM                 & PSNR  & SSIM                  \\
\hline
\hline
Bicubic~                    & 28.42 & 0.8104           & 26.00 & 0.7027            & 25.96 & 0.6675           & 23.14 & 0.6577               & 24.89 & 0.7866                \\
SRCNN~~& 30.48 & 0.8628           & 27.50 & 0.7513            & 26.90 & 0.7101           & 24.52 & 0.7221               & 27.58 & 0.8555                \\
FSRCNN~~                    & 30.72 & 0.8660           & 27.61 & 0.7550            & 26.98 & 0.7150           & 24.62 & 0.7280               & 27.90 & 0.8610                \\
VDSR~~ & 31.35 & 0.8830           & 28.02 & 0.7680            & 27.29 & 0.0726           & 25.18 & 0.7540               & 28.83 & 0.8870                \\
LapSRN~~                    & 31.54 & 0.8850           & 28.19 & 0.7720            & 27.32 & 0.7270           & 25.21 & 0.7560               & 29.09 & 0.8900                \\
MemNet~~                    & 31.74 & 0.8893           & 28.26 & 0.7723            & 27.40 & 0.7281           & 25.50 & 0.7630               & 29.42 & 0.8942                \\
SRMDNF~~                    & 31.96 & 0.8925           & 28.35 & 0.7787            & 27.49 & 0.7337           & 25.68 & 0.7731               & 30.09 & 0.9024                \\
D-DBPN~~                    & 32.47 & 0.8980           & 28.82 & 0.7860            & 27.72 & 0.7400           & 26.38 & 0.7946               & 30.91 & 0.9137                \\
RDN~~  & 32.47 & 0.8990           & 28.81 & 0.7871            & 27.72 & 0.7419           & 26.61 & 0.8028               & 31.00 & 0.9151                \\
SRFBN~~& 32.47 & 0.8983           & 28.81 & 0.7868            & 27.72 & 0.7409           & 26.60 & 0.8015               & 31.15 & 0.9160                \\
SAN  & 32.64 & 0.9003           & 28.92 & 0.7888            & 27.78 & 0.7436           & 26.79 & 0.8068               & 31.18 & 0.9169                \\

IGNN
& {32.57}
& {0.8998}
& {28.85}
& {0.7891}
& {27.77}
& {0.7434}
& {26.84}
& {0.8090}
& {31.28}
& {0.9182} \\

\hline
\hline
EDSR-S                 & 32.09 & 0.8938           & 28.58 & 0.7813            & 27.57 & 0.7357           & 26.04 & 0.7849               & 30.35 & 0.9067                \\
\textbf{+PCL} & \textbf{32.17} & \textbf{0.8948}           & \textbf{28.61} & \textbf{0.7825}            & \textbf{27.58} & \textbf{0.7365}           & \textbf{26.07} & \textbf{0.7863}               & \textbf{30.45} & \textbf{0.9078}                \\
\emph{Gains}    & \textcolor{green}{0.08}  & \textcolor{green}{0.0010}           & \textcolor{green}{0.03}  & \textcolor{green}{0.0012}            & \textcolor{green}{0.01}  & \textcolor{green}{0.0008}           & \textcolor{green}{0.03}  & \textcolor{green}{0.0014}               & \textcolor{green}{0.10}  & \textcolor{green}{0.0011}                \\
\hline
\hline
EDSR-L & 32.46 & 0.8968          & 28.80 & 0.7876            & 27.71 & 0.7420           & 26.64 & 0.8033               & 31.02 & 0.9148              \\
\textbf{+PCL} & \textbf{32.52} & \textbf{0.8993}           & \textbf{28.84} & \textbf{0.7881}            & \textbf{27.74} & \textbf{0.7428}          & \textbf{26.71} & \textbf{0.8056}              & \textbf{31.20} & \textbf{0.9171}               \\
\emph{Gains}     & \textcolor{green}{0.06}  & \textcolor{green}{0.0025}          & \textcolor{green}{0.04}  & \textcolor{green}{0.0005}            & \textcolor{green}{0.03}  & \textcolor{green}{0.0008}           & \textcolor{green}{0.06}  &\textcolor{green}{0.0019}               & \textcolor{green}{0.18}  & \textcolor{green}{0.0023}                \\
\hline
\hline
RCAN& 32.64 & 0.9002           & 28.85 & 0.7885            & 27.75 & 0.7432           & 26.75 & 0.8066               & 31.20 & 0.9170                \\
\textbf{+PCL}   & \textbf{32.70} &\textbf{0.9005}           & \textbf{28.89} & \textbf{0.7889}            & \textbf{27.78} & \textbf{0.7437}           & \textbf{26.84} & \textbf{0.8083}               & \textbf{31.41}& \textbf{0.9185}                \\

\emph{Gains}     & \textcolor{green}{0.06}  & \textcolor{green}{0.0003}           & \textcolor{green}{0.04}  & \textcolor{green}{0.0004}            & \textcolor{green}{0.03}  & \textcolor{green}{0.0005}           & \textcolor{green}{0.09}  & \textcolor{green}{0.0017}               & \textcolor{green}{0.21}  & \textcolor{green}{0.0015}                \\
\hline
\hline
HAN & 32.60 & 0.9000           & 28.90 & 0.7891            & 27.79 & 0.7440           & 26.84 & 0.8093               & 31.42 & 0.9174                \\
\textbf{+PCL}    & \textbf{32.68} & \textbf{0.9006}           & \textbf{28.92} & \textbf{0.7897}            & \textbf{27.80} & \textbf{0.7444}           & \textbf{26.95} & \textbf{0.8115}               & \textbf{31.44} & \textbf{0.9190}               \\

\emph{Gains}     & \textcolor{green}{0.08}  & \textcolor{green}{0.0006}           & \textcolor{green}{0.02}  & \textcolor{green}{0.0006}            & \textcolor{green}{0.01}  & \textcolor{green}{0.0004}           & \textcolor{green}{0.11}  & \textcolor{green}{0.0022}               & \textcolor{green}{0.02}  & \textcolor{green}{0.0016}  \\
\hline
\hline
NLSN& 32.58 & 0.9000           & 28.87 & 0.7890            & 27.76 & 0.7441           & 26.96 & 0.8115               & 31.27 & 0.9184                \\
\textbf{+PCL}   & \textbf{32.71} & \textbf{0.9012}          & \textbf{28.99} & \textbf{0.7907}            & \textbf{27.83} & \textbf{0.7455}           & \textbf{27.12} & \textbf{0.8151}              & \textbf{31.57} & \textbf{0.9213}   \\

\textit{Gains} & \textcolor{green}{0.13} &\textcolor{green}{0.0012} & \textcolor{green}{0.12} & \textcolor{green}{0.0017} & \textcolor{green}{0.07} & \textcolor{green}{0.0014} &\textcolor{green}{0.16} &\textcolor{green}{0.0036} &\textcolor{green}{0.30} & \textcolor{green}{0.0029} \\
\hline
\end{tabular}}
\end{center}
\end{table*}

In this paper, we propose an efficient contrastive learning framework for SISR as illustrated in \cref{fig:framework}. The discriminator and the SR network are trained alternately. When training the discriminator, we fix the SR network, whose parameters are shared from the previous step. When training the SR network, the embedding network is frozen and its parameters are the same as the discriminator of the previous step. 

For training, we crop patches of size $48\times48$ from LR image with the corresponding HR patches. We augment the training data with random horizontal flips and 90 rotations. Our model is trained by ADAM optimizer \cite{ADAM} with $\beta_1=0.9$ , $\beta_2=0.999$, and $\epsilon=10^{-8}$. The batch size we used is 16.

\begin{table*}[tbp]
\centering
\renewcommand{\arraystretch}{1}
\caption{Qaulitative comparison on test datasets.}
\label{tab:res_niqe_lpips}
\resizebox{0.75\textwidth}{!}{%
\begin{tabular}{|c|c|c|c|c|c|c|c|c|c|c|}
\hline

\multirow{2}{*}{Method}&\multicolumn{2}{c|}{Set5}         & \multicolumn{2}{c|}{Set14}          & \multicolumn{2}{c|}{B100}          & \multicolumn{2}{c|}{Urban100}      & \multicolumn{2}{c|}{Manga109}       \\
\cline{2-11}
 & NIQE& LPIPS  & NIQE& LPIPS  & NIQE& LPIPS  & NIQE& LPIPS  & NIQE& LPIPS\\
\hline
\hline
EDSR  & 7.136 & 0.1801 &6.128& 0.2688 & 6.351& 0.3023   & 5.435 & 0.2356  & 5.143  & 0.1364   \\

 \textbf{+PCL}  & \textbf{7.072} & \textbf{0.1790} & \textbf{6.021} & \textbf{0.2681} & \textbf{6.276} & \textbf{0.3017} & \textbf{5.413} & \textbf{0.2347} & \textbf{5.014} & \textbf{0.1343}  \\

\hline
RCAN&   7.275          & 0.1801         &6.215         & 0.2686         & \textbf{6.314}        & 0.3022          & 5.521          & 0.2324         & 5.225          & 0.1354           \\
 \textbf{+PCL}  & \textbf{6.964} & \textbf{0.1786} & \textbf{6.126} & \textbf{0.2681} & 6.389 & \textbf{0.3010} & \textbf{5.420} & \textbf{0.2314} & \textbf{5.043} & \textbf{0.1316}  \\
\hline
NLSN  &   7.050 & 0.1787         &6.164         & 0.2656       & 6.417        & 0.2975          & 5.586          & 0.2268       & 5.244          & 0.1311          \\
 \textbf{+PCL}   & \textbf{6.861} & \textbf{0.1766} & \textbf{6.214} &\textbf{0.2639} & \textbf{6.372} &\textbf{0.2971} & \textbf{5.414} &   \textbf{0.2256} &\textbf{5.090} & \textbf{0.1294} \\
\hline
\end{tabular}}
\end{table*}

\begin{figure*}[ht]
    \centering
    \includegraphics[width=\textwidth]{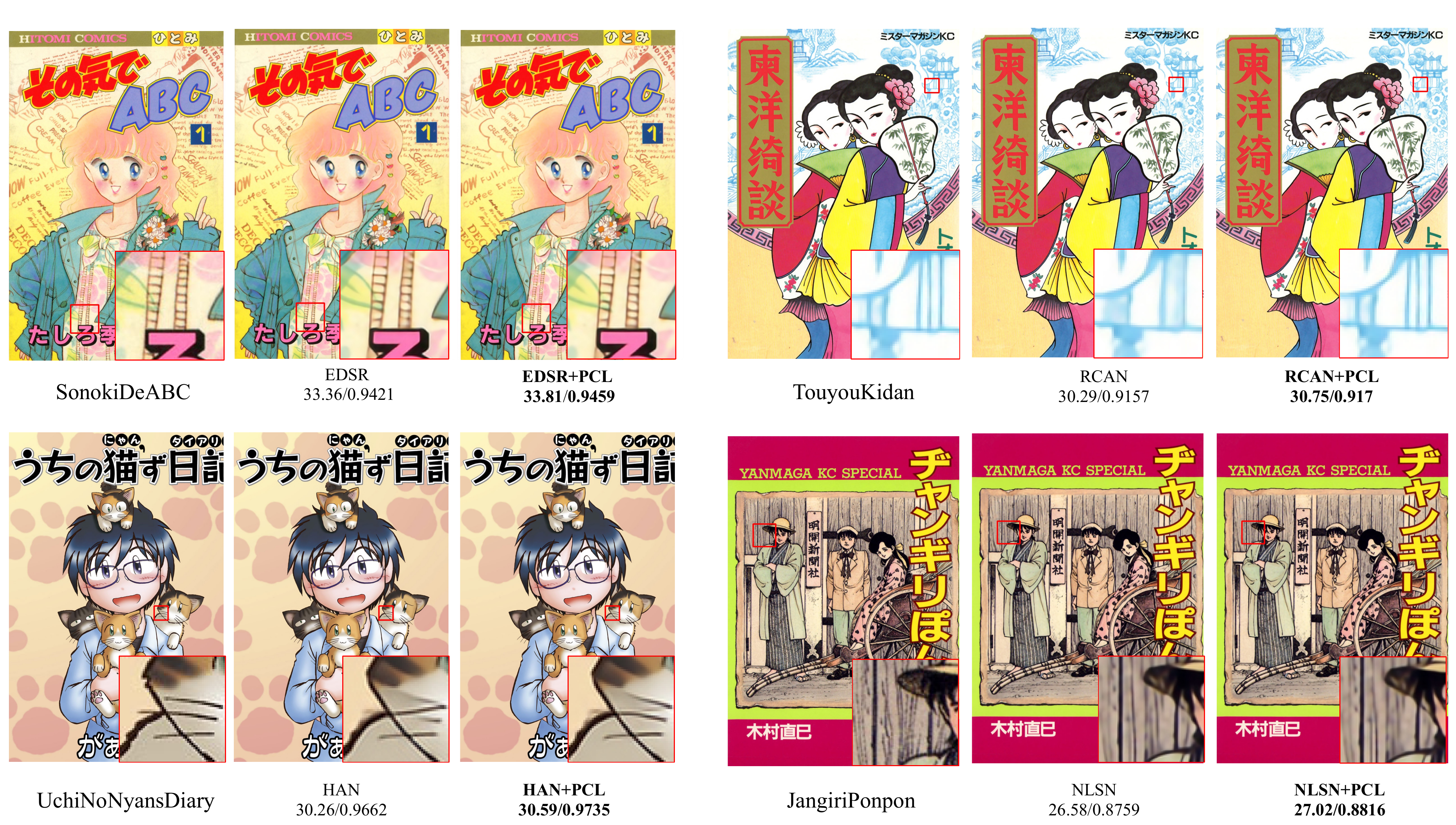}
    \caption{{\textbf{Visual comparison}. Comparisons of images with fine details. Results obtained by our PCL are in \textbf{bold} that contain more clear and more accurate reconstructed textures with fewer artifacts.}}
    \label{fig:patch_visual}
\end{figure*}

\section{Experiments\label{sec:experiments}}
\subsection{Experiment Setup}
\subsubsection{Datasets and Metrics.} Following \cite{EDSR,RCAN}, we use DIV2K dataset \cite{DIV2K} which contains 800 images for training and 100 images for evaluation. Datasets for testing include Set5 \cite{Set5}, Set14 \cite{Set14}, B100 \cite{B100}, Manga109 \cite{Manga109}, and Urban100 \cite{Urban100} with the up-scaling factor: $\times$4. For comparison, we measure PSNR and SSIM \cite{SSIM} on the Y channel of transformed YCbCr space.

\subsubsection{Comparison Methods.}. Our proposed contrastive learning framework for SISR task is generic and it can be applied to any existing method. We compare our method with the state-of-the-art (SoTA) methods. To evaluate our proposed method, we apply our propose ECL framework to existing benchmark methods and re-train them. Here we re-train EDSR\footnote{\url{https://github.com/sanghyun-son/EDSR-PyTorch}} \cite{EDSR}, RCAN\footnote{\url{https://github.com/yulunzhang/RCAN}} \cite{RCAN}, HAN\footnote{\url{https://github.com/wwlCape/HAN}} \cite{HAN}, and NLSN\footnote{\url{https://github.com/HarukiYqM/Non-Local-Sparse-Attention}} \cite{Non-local_sparse} with our contrastive learning framework. In addition, we also add some representative methods for comparison: SRCNN \cite{SRCNN}, FSRCNN \cite{FSRCNN}, VDSR \cite{Accurate_VDSR}, LapSRN \cite{LapSRn}, MemNet \cite{MemNet}, SRMDNF \cite{SRMDNF}, D-DBPN \cite{DBPN}, RDN \cite{RDN}, SRFBN \cite{SRFBN}, SAN \cite{SAN}, IGNN \cite{IGNN}, SwinIR \cite{SwinIR}. We use Pytorch \cite{pytorch} to implement our proposed approach and our code and all trained models can be achieved at our project page.

\begin{table*}[tbp!]
\begin{center}

\caption{Ablation study results. Here column $\mathcal{L}_{CL}$ presents types of embedding networks. VGG means adopting the pre-trained VGG network, and D presents our learnable discriminator. $H$ and $F$ are shorted for Haar and Fourier transformation respectively. $H$+D and $F$+D present that the embedding network takes the frequency transformed input .\label{tab:ablation}}
\renewcommand{\arraystretch}{1.2}
\resizebox{0.80\textwidth}{!}{%
\begin{tabular}{|c|c|c|c|c|c|c|c|c|c|c|c|c|}
\hline
\multirow{2}{*}{Config} & \multirow{2}{*}{$\mathcal{L}_{CL}$ } &\multirow{2}{*}{Aug}    & \multicolumn{2}{c|}{Set5} & \multicolumn{2}{c|}{Set14} & \multicolumn{2}{c|}{B100} & \multicolumn{2}{c|}{Urban100} & \multicolumn{2}{c|}{Manga109}  \\
\cline{4-13}
  &  && PSNR  & SSIM             & PSNR  & SSIM              & PSNR  & SSIM             & PSNR  & SSIM                 & PSNR  & SSIM                  \\
\hline
\hline
1 & VGG & \checkmark& 32.62&   0.8985&   28.84&  0.7875&27.73&0.7428&26.72&0.8059&31.12 & 0.9160\\
\hline
2 & D  &\checkmark &32.64&     0.8994  &28.86  &0.7889 & 27.76 & 0.7444 & 26.82 & 0.8082 & 31.33 & 0.9177\\
\hline 
3 & $H$+D & \checkmark&32.70 & 0.9005           & 28.89 & 0.7889            & 27.78 & 0.7437           & 26.84 & 0.8083               & 31.41 & 0.9185 \\
\hline
4 & $H$+D &  & 32.67 &  0.8999          & 28.87 &0.7885             & 27.76 & 0.7435          &26.83 & 0.8079               & 31.42 &0.9185 \\
\hline
5 & $F$+D & \checkmark & 32.66 &  0.8995          & 28.90 &0.7893             & 27.77 & 0.7434          &26.86 &0.8089               & 31.40 &0.9186 \\ 
\hline

\end{tabular}}
\end{center}
\end{table*}

\begin{figure}[h]
    \centering
    \includegraphics[width=0.45\textwidth]{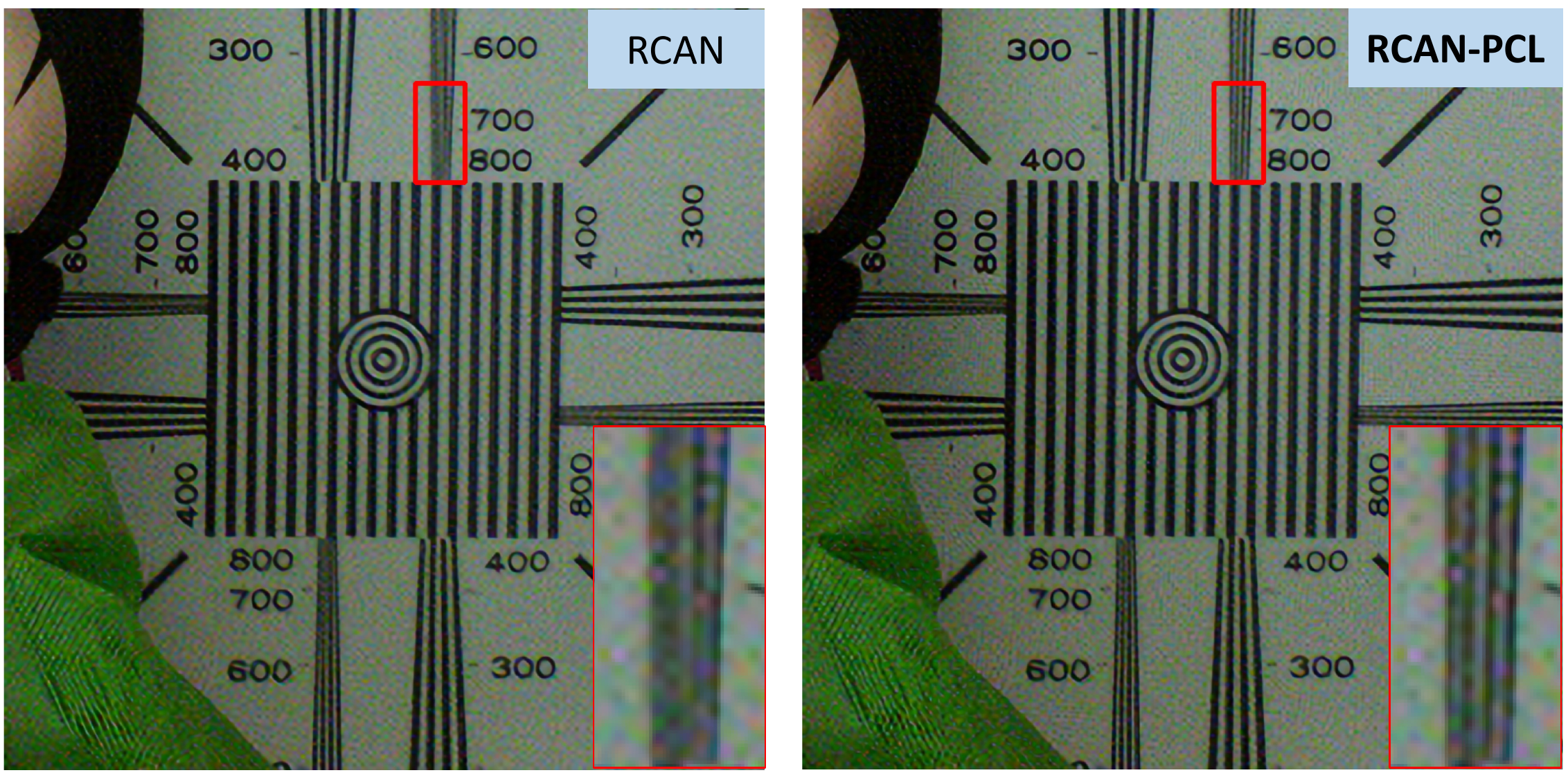}
    \caption{Results of the original and our re-trained RCAN on super-resolving real images from RealSRSet with scale factor 4. Compared to the RCAN (left), super-resolved images generated by RCAN-PCL (right) obtain clearer edges with fewer artifacts. Please zoom in for more details.}
    \label{fig:realsrset}
\end{figure}

\begin{table*}[!ht]
\begin{center}
\caption{Results of EDSR-S. The PCL-F presents that we re-train EDSR-S backbone with a fixed embedding network which is pre-trained with EDSR-L backbone.}
\label{tab:embedding}
\renewcommand{\arraystretch}{1.1}
\resizebox{0.75\textwidth}{!}{%
\begin{tabular}{|c|c|c|c|c|c|c|c|c|c|c|}
\hline
\multirow{2}{*}{Method}     & \multicolumn{2}{c|}{Set5} & \multicolumn{2}{c|}{Set14} & \multicolumn{2}{c|}{B100} & \multicolumn{2}{c|}{Urban100} & \multicolumn{2}{c|}{Manga109}  \\
\cline{2-11}
    & PSNR  & SSIM             & PSNR  & SSIM              & PSNR  & SSIM             & PSNR  & SSIM                 & PSNR  & SSIM                  \\
\hline
\hline
EDSR-S                 & 32.09 & 0.8938           & 28.58 & 0.7813            & 27.57 & 0.7357           & 26.04 & 0.7849               & 30.35 & 0.9067                \\
\hline
+PCL & 32.17 & 0.8948           & 28.61 & 0.7825            & 27.58 & 0.7365           & 26.07 & 0.7863               & 30.45 & 0.9078 \\
\hline
+PCL-F & 32.19 & 0.8948           & 28.62 & 0.7826            & 27.58 & 0.7365           & 26.09 & 0.7866               & 30.49 & 0.9079   \\         
\hline

\end{tabular}}
\end{center}
\end{table*}

\subsection{Main Results}
As summarized in \cref{tab:main_psnr_ssim}, we tabulate the quantitative results of different methods. The EDSR-S has 16 residual blocks and 64 channels, while EDSR-L is the large version and has 32 residual blocks and 256 channels. Compared to existing methods, one can find that all of our re-trained models surpass the original results on all the benchmark datasets. Specifically, compared with the NLSN model, our re-trained NLSN has a gain of 0.16 dB/0.0022 (mean value of PSNR/SSIM) on five benchmark datasets, and when compared with the SoTA SwinIR, re-trained NLSN can achieve comparable results on the term of PSNR among all benchmark datasets. In addition, re-trained HAN can obtain comparable performance with the origin NLSN. Our re-trained RCAN can achieve comparable performance with the newer SAN \cite{SAN} and especially it has superior performance on Set5, Urban100, and Manga109. From the \cref{tab:main_psnr_ssim}, one can find that the proposed PCL framework further improves the existing SR models, and when we take a closer look at the performance on the Manga109 dataset where the larger gains can be obtained by our PCL, it shows that the proposed PCL can further improve the generalization of the SR models as well.

\subsection{Qualitative Results}
To intuitively demonstrate the effectiveness of our proposed method, we present visual results of the official and re-trained models in \cref{fig:sr_visual}. One can find that the re-trained models produce clearer and more accurate textures than the official models by applying our contrastive learning constraint. Taking the images obtained by NLSN as an example, the re-trained models are able to effectively recover clear content in the presence of artifacts. Furthermore, the proposed PCL paradigm enables more accurate texture reconstruction. For example, the re-trained RCAN-PCL achieves accurate textures when the original RCAN fails to reconstruct, as shown in the second row of \cref{fig:sr_visual}. We provide additional visual comparisons in \cref{fig:patch_visual} with diverse contents from the cross-domain Manga109 test dataset. One can find that re-trained models always obtain more clear SR results. Furthermore, we present the SR results of both the original and re-trained RCAN model, applied to one typical low-quality image, which is characterized by real-world scenarios and presents unknown degradation. These results, as shown in \cref{fig:realsrset}, indicate that the RCAN-PCL model achieves clearer edges and fewer artifacts, revealing the robustness and generalization of the proposed PCL framework.

In addition, we provide a quantitative comparison of our approach on three large test datasets using the NIQE and LPIPS metrics, as presented in \cref{tab:res_niqe_lpips}. Our results indicate that SR models re-trained with the proposed PCL approach outperform the original models in most cases, revealing that the proposed PCL framework is effective at balancing distortion and perception.

\subsection{Ablation Study}

In this section, several ablation experiments are conducted to investigate different components in our proposed PCL-SR. We take the RCAN model as our baseline.

\subsubsection{Task-Generalized Embedding Network\label{sec:ablation}}
In contrast to some other methods that employ prior VGG network to build their contrastive loss \cite{contrastive_dehazing, wang2021towards}, our embedding network is learned with the SR network like the discriminator in the GAN framework to obtain the task-friendly embedding network.

\begin{table*}[!ht]
\begin{center}
\caption{Ablation studies about data augmentations on benchmark datasets (PSNR (dB)/SSIM/LPIPS).\label{tab:pos_neg_sample}}
\renewcommand{\arraystretch}{1.1}
\resizebox{0.70\textwidth}{!}{
\begin{tabular}{|c|c|c|c|c|c|c|c|c|c|c|}
\hline
  \multirow{2}{*}{Pos} &\multirow{2}{*}{Neg} & \multicolumn{3}{c|}{B100} & \multicolumn{3}{c|}{Urban100} & \multicolumn{3}{c|}{Manga109}  \\
\cline{3-11}
   &    & PSNR  & SSIM   &LPIPS & PSNR  & SSIM   &LPIPS & PSNR  & SSIM  &LPIPS   \\
\hline
\hline
 &       & 27.76 & 0.7435  &0.3011     &26.83 & 0.8079  &0.2318     &31.42 &0.9185&0.1319\\
\hline
 \checkmark & & 27.74 & 0.7433 &0.3007 & 26.82 & 0.8082 & 0.2307& 31.43 &0.9189 & 0.1318\\
\hline 
  & \checkmark&27.79 &0.7438   &0.3013 & 26.88 & 0.8094& 0.2316 & 31.40 & 0.9186 &0.1320 \\
\hline
 \checkmark  & \checkmark & 27.78 & 0.7437 & 0.3010 &26.84 & 0.8083  &0.2314 & 31.41 & 0.9185& 0.1316 \\
\hline 
\end{tabular}}
\end{center}
\vspace{-0.4cm}
\end{table*}

Here we analyze different types of the embedding network. Considering the prior VGG network is pre-trained in RGB space, for a fair comparison, we remove the wavelet transform and train our embedding network in RGB space as well, indicated as Config 2. Then we adopt the prior VGG-19 network to calculate our contrastive loss in RGB space, denoted as Config 1. The last is our full pipeline, where the wavelet transform is applied to highlight the difference in frequency space and this embedding network is learned with frequency map correspondingly, noted as Config 3. Results are reported in \cref{tab:ablation}. We can find that employing prior VGG network directly obtains even worse performance on benchmark datasets compared with baseline RCAN. Same with the observation in \cite{contrastive_dehazing,wang2021towards}, they propose their contrastive losses instead of the original normalized $\ell_{2}$-based contrastive loss because the original formulation cannot work properly. Since exploring a proper formulation for contrastive loss is beyond the scope of this paper, we adopt the original contrastive loss and extend it to a pixel-wise loss like \cite{cut}. Notably, our learned embedding network in Config 2 can achieve comparable performance with our basic contrastive loss, which shows that our embedding network is learned and becomes more proper for SISR task by distinguishing if the input is blurred. Finally, our full pipeline, employing the Haar transform, obtains the best performance.

In addition, when our embedding network is trained and finished, it is general to work with any SR backbones like the prior VGG. Here we re-train the EDSR-S backbone with fixed embedding network pre-trained with the EDSR-L backbone, noted with the suffix -F. Notably, when we apply our pre-trained embedding network, it is used in the same way as the pre-trained VGG network, and the results are shown in \cref{tab:embedding}. Some interesting observations can be obtained that the results of our PCL-F are slightly better. We think it is reasonable because the pre-trained embedding network based on the EDSR-L contains implicit knowledge to help the small backbone learn and using a fixed embedding network avoids the unstable problems in GAN-like training processing. Considering there are several great GAN-based real-world SR works such as \cite{real-esrgan}, discriminators trained contain rich prior about the backbone and degradation. We think it is meaningful to explore these task-friendly prior in pre-trained discriminators in feature works.

\subsubsection{{Sample Construction}}
As presented in \cref{tab:ablation}, we compare the performance of Configs 3 and 4 with and without the proposed task-valid sample generation, demonstrating its effectiveness. In addition, we provide a decoupled analysis of the impact of positive and negative samples in \cref{tab:pos_neg_sample}. Primarily, it is notable that the incorporation of 'hard' negative samples can contribute positively to obtaining credible results. This is anticipated to enhance the texture accuracy of the super-resolved results. Secondly, our findings suggest that refining positive samples tends to yield superior qualitative performance, thereby bolstering the generalization capabilities of the SR network. This improvement is particularly evident in the cross-domain test set Manga109, further underscoring the effectiveness of our approach.

\subsubsection{Spectrum Transformation}
To help the SR network pay more attention to the losing high-frequency information, our contrastive loss works with the embedded frequency features, and we take the Haar transform as the default. Here we take ablation studies to verify the impact of different types of wavelet transformation and results. Wavelet transformation is workable as discussed in \cref{sec:ablation}. Considering the Fourier transform is widely used as well, we adopt the FFT transform in Config 5, as shown in \cref{tab:ablation}. Comparing Config 3 and 5, we can find that utilizing the FFT transform can obtain comparable results as well, with better performance on Set14 and Urban100 while worse performance on Set5, B100, and Manga109. Since the transformed frequency map by Haar transform is only half the size of the original image which highly speeds up training processing and we take Haar transform as default. 

\begin{table}[h]
\begin{center}
\caption{The influence of hyper-parameter $\alpha$.\label{tab:alpha}}
\renewcommand{\arraystretch}{1.1}
\resizebox{0.4\textwidth}{!}{
\begin{tabular}{|c|c|c|c|c|c|c|}
\hline
$\alpha$   & 0.01   & 0.1  & 0.25   &0.5 & 1  & 5    \\
\hline
 PSNR    & 28.88 & 28.89  &28.84   &28.77 &28.4  &28.1 \\
\hline

\end{tabular}}
\end{center}
\end{table}
\subsubsection{Hyper-parameters}
To analyze the impact of the parameter $\alpha$, we conduct training of the RCAN-PCL model at scale 4, employing various $\alpha$ values. The results derived from the Set14 dataset are detailed in \cref{tab:alpha}. Observably, an increasing of the $\alpha$ value inversely affects the SR performance, leading to a decrease in effectiveness. In our experiments, we take $\alpha=0.1$ as the default value.

 {\subsection{Limitation and Discussion \label{sec:limitation}}}

 {In this paper, we leverage contrastive learning approaches for low-level SISR tasks. While our proposed PCL paradigm can enhance existing SR models, we also identify some challenges. Firstly, in the proposed PCL, hard negative samples require a well-designed prior. It is not clear whether a potential metric can guide and improve sample generation. Secondly, we adopt a GAN-like framework to train the task-specific embedding network, which introduces some minor uncertainty during training. As shown in Table \ref{tab:embedding}, a pre-trained embedding network can further improve performance. This observation is interesting and warrants further investigation in future work.}

  {Our findings contribute to the development of contrastive learning approaches for low-level SR tasks. However, there are still challenges that need to be addressed in the future. We believe that improving task-friendly sample generation for image restoration tasks is a promising research direction. Moreover, as presented in Table \ref{tab:embedding}, embedding networks translated from pre-trained SR models can achieve further improvement. For the low-level community, we suggest exploring how to extract more meaningful information or regularization from pre-trained models.}

\section{Conclusion \label{sec:conclusion}}
In this study, we investigate the contrastive learning approaches in low-level tasks and propose a practical contrastive learning framework for low-level SISR task, named PCL-SR. We take the contrastive learning paradigm into low-level SISR task from two perspectives: sample selection and feature embedding. Thus valid data augmentations and task-friendly embedding networks are proposed. In contrast to the previous methods \cite{han2021single,contrastive_dehazing,chen2021unpaired} that build contrastive samples with low-quality input and high-quality ground truth and employ the pre-trained and fixed embedding network, we generate the negative and positive samples by our proposed task valid augmentations, and to well captured the degradation-related difference, we employ an adaptive and learnable embedding network to obtain the task-friendly features. In experiments, some representative SR models are re-trained by the proposed PCL. The re-trained models achieve superior performance compared to the origin among all the test datasets, and detailed ablation studies are conducted to analyze the impact of different components. Furthermore, our PCL paradigm is generic and can be applied to many other image restoration tasks, and we will explore contrastive learning approaches for more low-level tasks in the future. Besides, as we summarized in \cref{tab:embedding}, expanding our PCL to explore and utilize the implicit information in pre-trained models is another promising direction especially since there are more and more huge pre-trained models proposed nowadays.

\bibliographystyle{unsrt}
\bibliography{TNNLS}

\end{document}